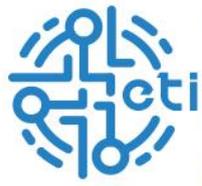
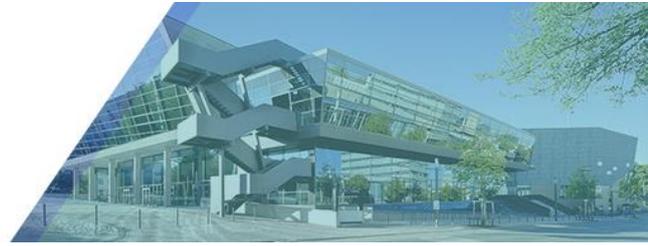

# Review of deep learning in healthcare

**Hasan Hejbari Zargar 1, Saha Hejbari Zargar 2, 3, Raziye Mehri 2, 3,**

1. Islamic Azad University of Ardabil, Ardabil, Iran
2. Deputy of Research and Technology, Ardabil University of Medical Sciences, Ardabil, Iran
3. Department of Community Medicine, Faculty of Medicine, Ardabil University of Medical Science, Ardabil, Iran

**Abstract**

Given the growing complexity of healthcare data over the last several years, using machine learning techniques like Deep Neural Network (DNN) models has gained increased appeal. In order to extract hidden patterns and other valuable information from the huge quantity of health data, which traditional analytics are unable to do in a reasonable length of time, machine learning (ML) techniques are used. Deep Learning (DL) algorithms in particular have been shown as potential approaches to pattern identification in healthcare systems. This thought has led to the contribution of this research, which examines deep learning methods used in healthcare systems via an examination of cutting-edge network designs, applications, and market trends. To connect deep learning methodologies and human healthcare interpretability, the initial objective is to provide in-depth insight into the deployment of deep learning models in healthcare solutions. And last, to outline the current unresolved issues and potential directions.

**Keywords:**  Machine learning Deep neural network Healthcare applications Diagnostics tools Health data analytics

# 1. Introduction

A new age in health care is rapidly approaching, one in which the wealth of biological data will play an increasingly significant role. By considering various aspects of a patient's data, such as variability in molecular traits, environment, electronic health records (EHRs), and lifestyle, precision medicine, for instance, aims to "ensure that the right treatment is delivered to the right patient at the right time." The abundance of biomedical data presents both enormous potential and difficulties for health care research. To build trustworthy medical solutions based on data-driven techniques and machine learning, a major difficulty is examining the relationships among all the many bits of information in these data sets. Previous research has attempted to combine various data sources in order to create collaborative knowledge bases that can be utilized for discovery and predictive analysis [19].

In the medical field, which places enormous demands on human life, the healthcare service system is crucial. Healthcare professionals in developing nations are using intelligent technology, such as artificial intelligence (AI) and machine learning methods, to advance their professions. Healthcare innovation has influenced research on intelligent healthcare systems that are oriented on people. AI technologies have an impact on how intensive care and administrative tasks are developed in hospitals and clinics. Since 2019, Jafar Abdollahi has conducted extensive research in the area of disease diagnosis with artificial intelligence. According to his findings, artificial intelligence, including machine learning and deep learning, has been successfully used in medical image and healthcare analysis for diseases like Diabetes [1, 10, 18], Breast Cancer [2, 11], Healthcare System [3], Forecasting [4, Stock Market [5], Stroke [6, COVID-19 [7], Types of the Epidemic [8, Medicinal Plants [9], and Heart [12].

Deep learning techniques haven't, however, been thoroughly examined for a wide variety of medical issues that can benefit from their capabilities. Deep learning has numerous features that might be used in the healthcare industry, including its better performance, end-to-end learning model with integrated feature learning, capacity to handle complicated and multi-modality data, and more. The deep learning research community as a whole needs to address a number of issues related to the characteristics of health care data (i.e., sparse, noisy, heterogeneous, and time-dependent), as well as the need for improved techniques and tools that allow deep learning to interface with clinical decision support workflows.

In this article, we cover current and upcoming deep learning applications in medicine, emphasizing the crucial elements to have a substantial influence on health care. We do not want to provide a thorough foundation on the technical aspects or widespread applications of deep learning. As a result, in the sections that follow, we'll give a brief overview of the general deep learning framework, examine some of its applications in the medical field, and talk about the advantages, drawbacks, and potential uses of these techniques in the context of precision medicine and next-generation health care. The taxonomy of popular deep learning architectures for HCS data analysis is shown in Fig. 1, along with a few HCS applications, particularly one for illness detection [23].

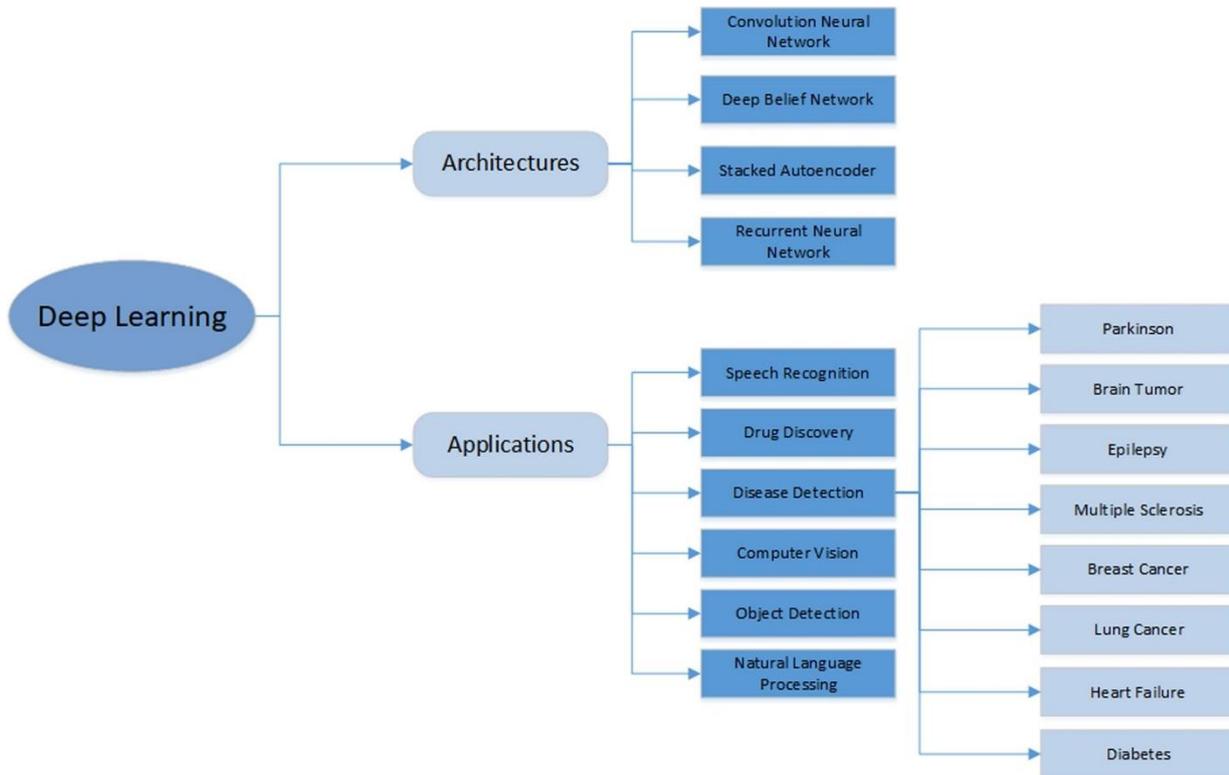

**Fig 1.** Application and Architectures of Deep Learning.

## 2. Advantages of DL and ML in Medical Healthcare Data

The deep learning and machine learning are evolutionary changes in various fields such as industry, companies, schools, colleges, and healthcare systems, and we can say that more changes are seen in the medical line by providing many types of online and offline facilities. Deep learning plays an important role in the detection of cancer cells automatically. The ML can solve multiple tasks but it needs human beings, while the DL performs alone or automatically using machine learning. Deep learning solves the whole problem unlike ML automatically. Deep learning is more beneficial for elders, coma patients, and cardiac disease diagnosis, especially in case of children [22]. The application of AI and other approaches is shown in Figure 2.

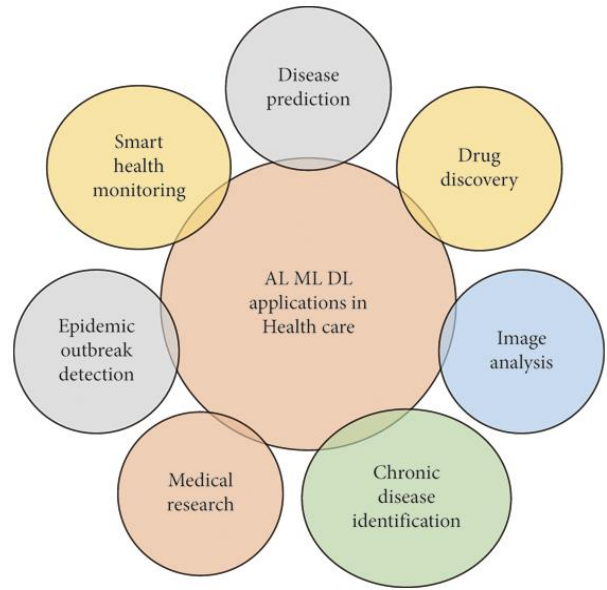

Figure 2. Applications of ML and DL in the healthcare system.

## 2.1. Deep learning framework

A general-purpose artificial intelligence technique called machine learning may infer correlations from data without first defining them. The key selling point is the capacity to generate predictive models without making firm assumptions about the underlying processes, which are often unidentified or inadequately characterized. Data harmonization, representation learning, model fitting, and assessment make up the usual machine learning pipeline. For many years, building a machine learning system needed meticulous engineering and subject-matter knowledge to convert the raw data into an appropriate internal representation from which the learning subsystem, often a classifier, could find patterns in the data set. Conventional methods can only handle natural data in its raw form since they only involve one, often linear, modification of the input space [19].

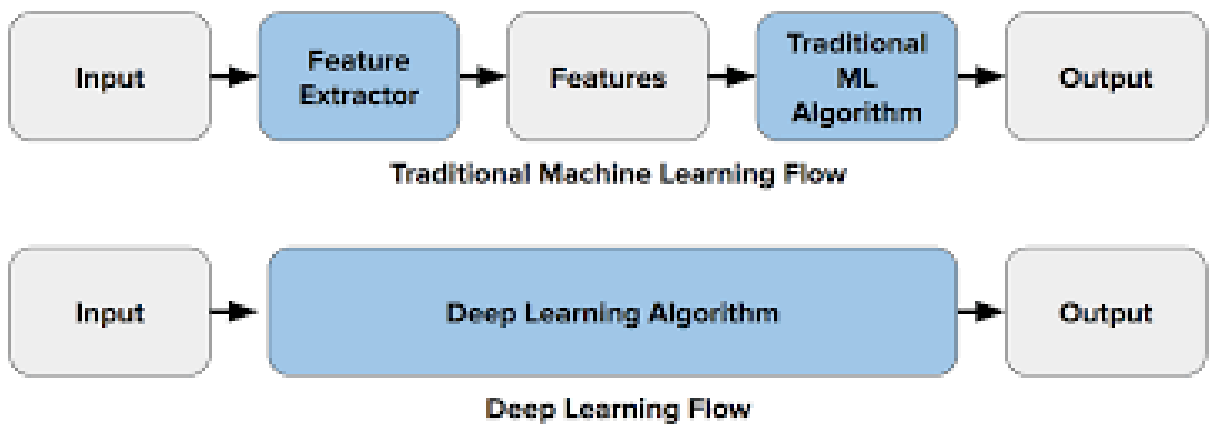

**Fig 3.** Deep learning framework

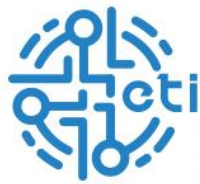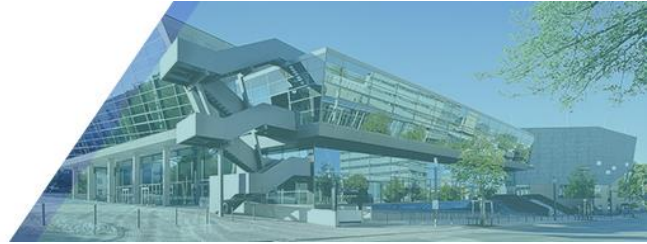

### A. Fully Connected Neural Networks (FCNNS)

The FCNNs consist just of neurons and layers of models, where each layer's input is connected to each neuron in the one below. The most straightforward way to understand neurons is to conceive of them as linear regression models, where each neuron uses input data (x), weights (y), and bias (z) to create an output (y). If the result after the activation function is zero, the neuron will be eliminated for that particular instance. The neuron's value must pass through the activation function. In any other case, the value will be sent to the network's next tier. Depending on the kind of activation function utilized, the output value changes. For instance, the output of the sigmoid activation function will have an s-shaped curve and range between 0 and 1. Rectified Linear Unit (ReLU) activation function is another example. For values greater than 0, it has a linear form, while for values less than or equal to 0, it has a 0 [20].

### B. Convolutional Neural Networks (CNN)

The most well-known and often used deep learning architecture for image data is CNN. CNNs are layers that use convolutional techniques to certain inputs to create specific outputs. In the convolution process, filters are utilized as a sliding window to scan over all input areas and create a feature map. To reduce the number of created features and subsequently the computation, pooling layers with convolution layers are utilized in the downsampling process. The output often has a similar form to the input, which may be a tensor in one, two, or three dimensions. Many other CNN designs have been created over the years; among of the most well-known ones are LaNet, ResNet, VGGNET, EfficientNet, and others. CNNs are employed in numerous machines learning applications, including video processing, natural language processing (NLP), and time series prediction, in addition to being the most widely used approach for image processing. In the last few years, CNN has also grown in prominence in BDL. They are utilized in a variety of applications, including medical imaging [20], text detection, genomic research, and others.

### C. Recurrent Neural Networks (RNNS)

For deep sequential learning, which is utilized for sequence or streaming data, such as video, audio, and time-series prediction, RNNs are one of the most often employed architectures [32]. The neurons in the same layer are connected in these networks by recurrent "cycle" linkages. RNNs include memory cells that provide the model the ability to retain data from the past, which is crucial for predicting future events. In order to anticipate the output of the network, RNNs must maintain a state of the past and current input. To put it another way, technically, computing the outcome requires knowledge of the status of every prior input. The vanishing gradient descent issue arises when the sequence becomes longer, making the starting weight unreachable and lowering the performance of the RNNs. As a result, many architectural designs have been created to address this problem; one such design is the Long Short-Term Memory (LSTM) concept. Throughout the learning process, the LSTM is advised for managing each memory cell for both state and output values [20].

## 2.2. Challenges and opportunities

Despite the encouraging outcomes produced by deep architectures, there are still a number of issues that need to be resolved before deep learning in healthcare may be used clinically. We draw special attention to the following significant problems. [19]:

- **Data volume**: Deep learning describes a group of computer models that are very labor-intensive. Fully linked multi-layer neural networks are one common example, where several network parameters must be accurately predicted. The availability of a vast quantity of data serves as the foundation for achieving this aim. Despite the fact that there are no strict rules regarding the minimum number of training documents, it is often recommended to use at least 10 times as many samples as parameters in the network. This is also one of the explanations for why deep learning is so effective in fields like computer vision, voice, and natural language processing where massive amounts of data are amenable to easy collection. Although there are only 7.5 billion people in the globe (as of September 2016), a large portion of them do not have access to basic healthcare, making health care a separate field. As a result, we are unable to get enough patients to build a thorough deep learning model. Additionally, comprehending illnesses and their variations is a lot more difficult than other tasks, like voice or picture recognition. Therefore, in comparison to other media, the quantity of medical data required to train an efficient and reliable deep learning model would be much more from a big data viewpoint [19].
- **Data quality**: Health care data are very varied, confusing, noisy, and incomplete, in contrast to other fields where the data are clean and well-structured. It is difficult to train an effective deep learning model with such large and diverse data sets and must take into account a number of factors, including data sparsity, redundancy, and missing values [19].
- **Temporality**: The illnesses constantly advance and change in a nondeterministic fashion across time. However, a lot of current deep learning models—including some that have been suggested for the medical field—assume static vector-based inputs, which do not naturally handle the time aspect. The creation of unique solutions will be necessary for designing deep learning methods that can handle temporal health care data [19].

Domain-level complexity the issues in bioscience and healthcare are trickier than in other application sectors (like voice and image analysis). Due to the extreme heterogeneity of the illnesses, it is still unclear exactly what causes most of them and how they develop. Furthermore, in a real-world clinical setting, the number of patients is often constrained, therefore we are unable to request an unlimited number of patients.

- **Interpretability**: Deep learning models are sometimes seen as "black boxes," despite the fact that they have proven effective in a number of application fields. While this might not be a problem in other, more deterministic domains like image annotation (where the end user can objectively validate the tags assigned to the images), in the field of health care, it is important to understand not only how well the algorithms perform quantitatively but also why they do so. In fact, such model interpretability (i.e. revealing which phenotypes are driving the predictions) is essential for persuading the medical community to take the actions suggested by the predictive system (e.g. prescribing a specific medication, possibly having a high risk of contracting a specific disease) [19].

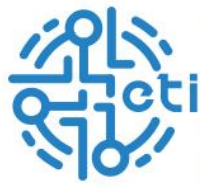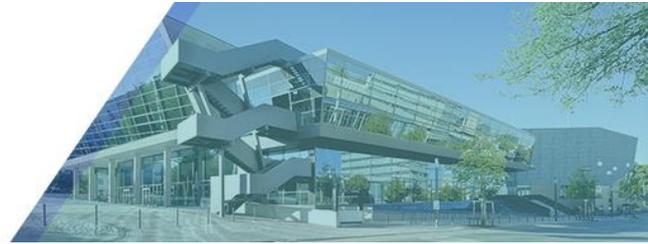

## 3. Treatment Applications of ML

**a) Image interpretation**

As was previously said, medical pictures are often employed in everyday clinical practice, and skilled radiologists and doctors analyze and interpret these images. They generate textual radiology reports on each bodily part that was evaluated in the performed research in order to describe the results about the pictures being reviewed. Writing such reports, however, may be particularly difficult in certain situations, such as for radiologists with less training or for healthcare practitioners in remote places where the standard of treatment is subpar. The process of creating high-quality reports, on the other hand, may be laborious and time-consuming for expert radiologists and pathologists, which can be made worse by the many patients that come in each day. As a result, several academics have made an effort to solve this issue utilizing natural language processing (NLP) and machine learning (ML) approaches. [21] proposes a technique for annotating clinical radiological data using natural language processing. For the automated labeling and description of medical pictures, a multi-task machine learning (ML) based architecture is suggested [21]. A similar work [21] presents an end-to-end architecture for thorax illness categorization and reporting in chest X-rays that was constructed using the combination of CNN and RNN. A unique multi-modal model for automated report production is created in [21] using an LSTM network and CNN.

**b) Real-time health monitoring using ML**

Monitoring critical patients in real time is essential and a fundamental step in the healing process. People are becoming more interested in continuous health monitoring via smartphones, IoT sensors, and wearable technology. A wearable gadget and a smartphone are often used to gather health data for continuous health monitoring, which is subsequently sent to the cloud for ML/DL analysis. The results are then sent back to the device so it may take the necessary action(s). For instance, [21] presents a framework with a comparable system design. The system is created by fusing mobile and cloud for PPG signal-based heart rate monitoring. Similar to this, [21] presents a study of several ML approaches for human activity identification with application to remote patient monitoring utilizing wearable technology. We go through several of the privacy and security issues raised by the sharing of health data with clouds for further analysis in the next section.

## 4. Conclusion

Deep learning techniques are potent tools that enhance conventional machine learning and enable computers to learn from the data in order to find ways to develop smarter applications. A variety of applications, particularly in computer vision and natural language processing, have already exploited these methods. The findings from every study that has been published in the literature demonstrate how deep learning can also be used to analyze health care data. In fact, using multi-layer neural networks to handle medical data boosted the prediction capacity for a number of particular applications across several clinical domains. Furthermore, given the emphasis on representation learning and not just classification accuracy, deep architectures have the ability to integrate various data sets across disparate data types and provide higher generalization due of their hierarchical learning structure [24-34].

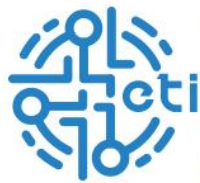
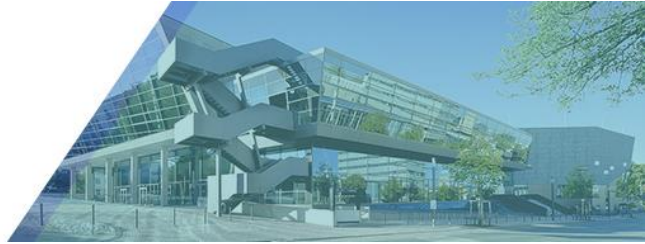

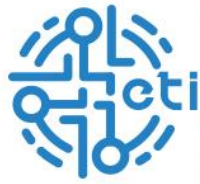
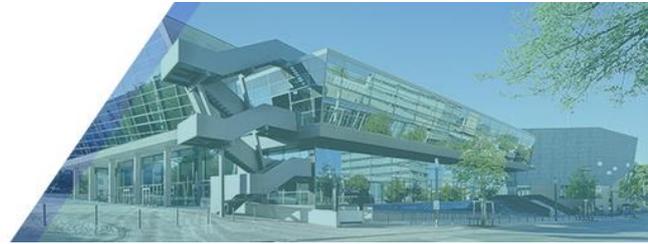